\pdfoutput=1
\documentclass{article}

\nocite{*} %
\usepackage[preprint]{neurips_2026}

\usepackage[utf8]{inputenc} 
\usepackage[T1]{fontenc}    
\usepackage{hyperref}       
\usepackage{url}            
\usepackage{booktabs}       
\usepackage{amsfonts}       
\usepackage{amsmath}        
\usepackage{nicefrac}       
\usepackage{microtype}      
\usepackage{xcolor}         
\usepackage{graphicx}       
\usepackage{float}          
\usepackage[ruled,vlined,linesnumbered]{algorithm2e}
\usepackage{multirow}
\usepackage{tabularx}
\usepackage{geometry}
\usepackage{colortbl}
\usepackage{graphicx}

\title{Logit Distillation on Manifolds: Mapping by Learning}

\author{%
  Yiru Yang\thanks{Equal contribution.} \\
  University of Zurich \\
  \texttt{yiru.yang@uzh.ch} \\
  \And
  Junling Wang\footnotemark[1] \\
  ETH Zurich \\
  \texttt{junling.wang@ethz.ch} \\
  \And
  Nishant Kumar Singh\footnotemark[1] \\
  University of Zurich \\
  \texttt{nishantkumar.singh@uzh.ch} \\
  \AND
  Luohong Wu \\
  University of Zurich \\
  \texttt{luohong.wu@uzh.ch} \\
  \And
  Haoran Yan \\
  Deutsche Bank Securities\\
  \texttt{haoran.yan@db.com} \\
}

\begin{document}

\maketitle

\begin{abstract}
A simple way to improve the performance of almost any machine learning model is not to train a single but several models with diverse algorithms which will make slightly distinct kinds of predictions and errors on the same data, and thus improve the average predictions and robustness. However, making predictions using a whole ensemble of models is cumbersome and computationally too expensive to allow deployment to a large number of users, especially if the models are large neural nets. In response to this, we introduce a layer and point wise projection mapping, which maps student and teacher representations into an aligned high-dimensional embedding space during training process. The proposed approach combined with LoRA injection reduces the student model trainable parameters to less than 1\% of the teacher model, while significantly improving word error rate (WER) compared to other distillation methods, as demonstrated in ablation studies. Unlike a mixture of experts, our method can be trained rapidly and in parallel. 
\end{abstract}

\section{Introduction}

Large foundation models have demonstrated that scaling alone is insufficient; the underlying representation geometry also plays a critical role in efficient knowledge transfer and generalization. In large language models and speech foundation models, knowledge distillation has become a standard paradigm for compressing large teacher networks into smaller student architectures while preserving downstream performance. Classical logit distillation \citep{hinton2015distilling} formulates this process as matching softened output distributions through KL divergence. Empirically, this strategy enables substantial reductions in inference cost and memory footprint while maintaining competitive predictive accuracy.

However, existing distillation methods implicitly assume that teacher and student representations lie in a compatible Euclidean space. Under the high-temperature approximation of \citep{hinton2015distilling}, minimizing the distillation objective becomes equivalent to point-wise quadratic matching between logits. Although effective in practice, this formulation treats representations as isolated vectors and ignores the intrinsic geometric structure of latent spaces. As modern foundation models become increasingly deep and heterogeneous, such Euclidean alignment becomes insufficient for preserving higher-order relational information across layers and modalities.

Recent advances in representation learning suggest that learning dynamics are fundamentally geometry-dependent. Information geometry interprets probability distributions as points on a statistical manifold equipped with the Fisher information metric, while recent works on hyperspherical and normalized transformers demonstrate that constraining optimization trajectories to structured manifolds can significantly improve convergence and stability. At the same time, geometry-aware learning paradigms in contrastive learning and self-supervised learning emphasize preserving relational structure rather than performing naive coordinate-level matching. These observations motivate revisiting knowledge distillation from a geometric perspective.

In this work, we propose \textbf{Riemann-Constrained Logit Distillation (RC)}, a geometry-aware distillation framework that lifts classical logit alignment from Euclidean space to a learned Riemannian manifold. Instead of directly matching logits or hidden vectors, we model intermediate representations as subspaces and align them through learnable projection mappings. Concretely, we introduce a layer-wise projection module $\phi_i$, which maps student representations into the ambient feature space of the teacher and induces a pullback Riemannian metric:
\[
g_\phi = J_\phi^\top J_\phi,
\]
where $J_\phi$ denotes the Jacobian of the projection function. This formulation allows the geometry of the alignment space itself to be learned during training.

To preserve structural information beyond point-wise embeddings, we further formulate representation alignment on the Grassmann manifold. The student and teacher hidden states are interpreted as subspaces rather than individual feature vectors, and geometric consistency is enforced through subspace projection operators. Unlike conventional distillation, which only transfers output probabilities, our method explicitly transfers latent geometric structure across network depths.

We instantiate the proposed framework on multilingual automatic speech recognition (ASR) using Whisper-based teacher-student architectures. The teacher model is Whisper Large-v3, while the student model is Whisper Medium with frozen backbone parameters. Adaptation is restricted to lightweight LoRA modules and the proposed geometry-aware projection layers, resulting in less than $0.1\%$ trainable parameters relative to the teacher. Despite this extreme compression regime, the proposed method substantially improves word error rate (WER) compared to standard LoRA-based and KL-based distillation baselines.

Our formulation is inspired by the broader observation that successful generative and autoregressive systems increasingly rely on structural rather than purely local alignment. Similar to how Visual AutoRegressive Modeling (VAR) reformulates image generation from rasterized next-token prediction into hierarchical next-scale prediction to preserve spatial structure and improve scalability :contentReference[oaicite:0]{index=0}, our work reformulates knowledge distillation as geometry-aware structural alignment instead of flat Euclidean point matching. The central idea is that preserving the organization of representations is often more important than matching individual coordinates.

The main contributions of this work are summarized as follows:
\begin{enumerate}
    \item Reinterpret knowledge distillation as optimization on a learned Riemannian manifold rather than Euclidean logit matching.
    
    \item Introduce a geometry-aware projection framework that aligns student and teacher representations through layer-wise learnable mappings and Grassmann subspace alignment.
    
    \item Propose a parameter-efficient distillation pipeline combining frozen backbones, LoRA adaptation, and Riemannian projection learning, achieving over $99.79\%$ training compression.
    
    \item Demonstrate the effectiveness of the proposed framework on multilingual ASR using Whisper models, showing significant improvements over classical distillation baselines under extreme parameter constraints.
\end{enumerate}

\section{Related Work}

\subsection{Knowledge Distillation}

\textbf{KL-based logit distillation.}
Knowledge distillation compresses the behavior of a high-capacity teacher model into a smaller student model by training the student to match the teacher's softened output distribution \citep{ghani2013proceedings,hinton2015distilling}. 
Compared with hard labels, soft targets expose ``dark knowledge'' about inter-class similarities and teacher uncertainty, providing a richer supervision signal for model compression \citep{hinton2015distilling}. 
Classical distillation is commonly formulated as minimizing the KL divergence between teacher and student predictive distributions under temperature-scaled softmax outputs. 
Under a high-temperature approximation, this objective is closely related to matching teacher and student logits with a quadratic objective \citep{hinton2015distilling,liu2018teacher}. 
Recent work has revisited logit-level distillation by studying the role of logit scale, normalization, and teacher-student capacity mismatch \citep{sun2024logit}. 
Despite these advances, logit distillation remains primarily an output-level alignment objective, and it does not explicitly model the geometry of the latent representation space. 
In contrast, our work preserves the original KL-based distillation signal while extending teacher-student alignment to a learned geometric representation space.

\textbf{Intermediate representation alignment.}
Beyond output-level supervision, a large body of work distills intermediate representations from teacher networks. 
FitNets introduced hidden-layer hints, where a student is trained to predict intermediate teacher representations through a learned regressor \citep{romero2014fitnets}. 
Attention Transfer further showed that matching teacher attention maps can improve student training \citep{zagoruyko2016paying}. 
Other methods preserve relational structure by aligning pairwise activation similarities or sample relationships rather than individual features alone \citep{tung2019similarity,park2019relational}. 
Contrastive Representation Distillation formulates representation transfer as a contrastive learning problem, encouraging the student to capture more information from teacher representations than is provided by output distributions alone \citep{tian2019contrastive}. 
These methods demonstrate that intermediate representations contain useful transfer signals, but they typically operate through Euclidean feature regression, attention-map matching, or similarity objectives defined over vector embeddings. 
Our method differs by treating hidden representations as subspaces and aligning them on a learned Riemannian manifold, enabling geometry-aware transfer beyond point-wise feature matching.

\subsection{Geometry-Aware Representation Learning}

\textbf{Information geometry and manifold learning.}
Information geometry studies probability distributions as points on statistical manifolds equipped with intrinsic metrics such as the Fisher information metric \citep{amari2000methods,amari2016information}. 
This perspective has influenced optimization and representation learning by emphasizing that the choice of geometry affects distances, gradients, and learning trajectories. 
More broadly, Riemannian optimization provides tools for learning under non-Euclidean constraints, including optimization on Stiefel and Grassmann manifolds \citep{absil2008optimization}. 
Grassmannian methods are especially relevant when data are better represented as linear subspaces rather than individual vectors \citep{hamm2008grassmann,huang2015projection}. 
These perspectives suggest that teacher-student alignment should preserve intrinsic geometric structure, rather than only coordinate-level similarity in a fixed Euclidean space.

\textbf{Geometry-aware optimization and representation learning.}
Non-Euclidean representation learning has shown that choosing an appropriate geometry can improve the capacity and inductive bias of learned embeddings. 
For example, Poincar\'e embeddings use hyperbolic geometry to represent hierarchical structure more effectively than Euclidean embeddings \citep{nickel2017poincare}. 
Contrastive learning objectives such as InfoNCE also preserve relational structure through similarity-based supervision rather than direct coordinate matching \citep{oord2018representation}. 
Recent normalized architectures such as nGPT further show that constraining embeddings, hidden states, and parameter vectors to hyperspherical geometry can improve training efficiency and representation stability \citep{loshchilov2025ngpt}. 
Inspired by these observations, our work formulates distillation as geometry-aware representation alignment: a learnable projection induces the metric of the alignment space, and student-teacher representations are matched as structured subspaces rather than isolated Euclidean vectors.

\section{Methodology}
\label{sec:method}

\subsection{Projection}

\begin{figure}[t]
    \centering
    \includegraphics[width=0.88\linewidth]{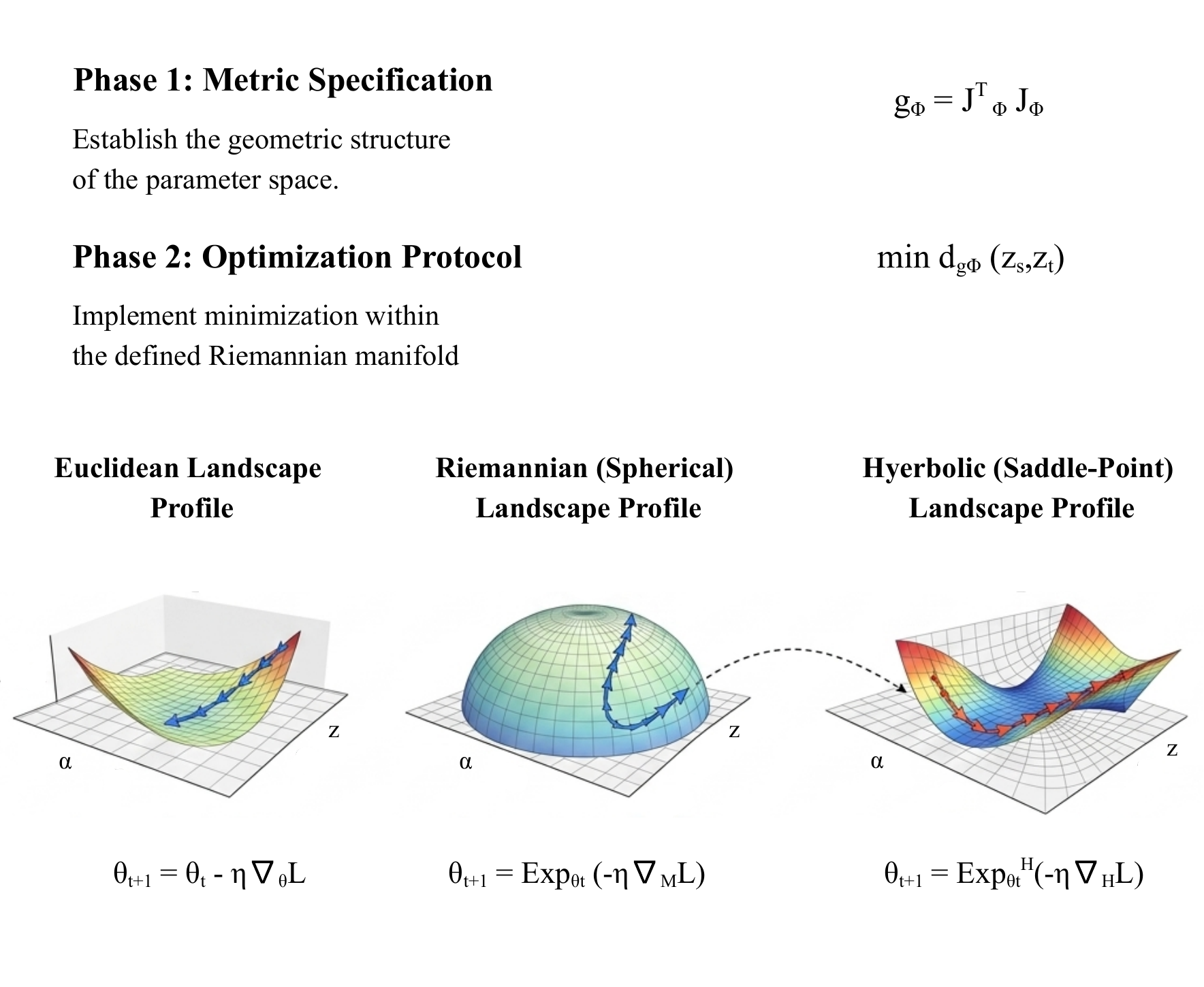}
    \caption{\textbf{Transformation of loss landscape geometry.} Standard optimisation assumes a flat Euclidean space (left). By defining the metric tensor $g_{\phi}$ and minimising distances under the induced geometry, optimisation moves into non-Euclidean spaces. The resulting trajectories are illustrated for Riemannian (center) and hyperbolic (right) geometries, showing how learned geometric mappings reshape optimisation paths and improve stability.}
    \label{fig:figure1}
\end{figure}

Instead of aligning individual embeddings on a sphere, we model encoder representations as subspaces and perform alignment on the Grassmann manifold: (i) the \textit{sphere} represents "vector alignment", while (ii) the \textit{Grassmannian} represents "subspace alignment". For interpretation, classical distillation operates in a flat euclidean space, but our formulation lifts it to a learned Riemannian manifold; thus making knowledge transfer geometry-aware alignment rather than pointwise matching.

\paragraph{Subspace Modeling and Alignment.}
We model encoder representations as subspaces rather than individual vectors. Let $h_s \in \mathbb{R}^{T \times d_s}, \quad h_t \in \mathbb{R}^{T \times d_t}$ denote the final encoder hidden states of the student and teacher networks, respectively, where $d_s \neq d_t$.

\paragraph{Step 1: Projection.}
To resolve the dimensional mismatch between student and teacher representations, we introduce a learnable linear projection:
\[
h_s \to W h_s, \quad W \in \mathbb{R}^{d_t \times d_s}
\]
This projection maps the student representations into the ambient space of the teacher, enabling subsequent geometric comparison.

\paragraph{Step 2: Orthogonalization.}
Raw hidden representations are not uniquely defined due to basis ambiguity, i.e., multiple matrices can span the same subspace. To obtain a canonical representation, we orthogonalize both projected student features and teacher features:
\[
U_s = \text{orth}(W h_s), \quad U_t = \text{orth}(h_t)
\]
where $U_s, U_t \in \mathbb{R}^{d_t \times k}$ are orthonormal bases satisfying:
\[
U_s^T U_s = I, \quad U_t^T U_t = I
\]

This step removes arbitrary linear transformations and ensures that comparisons are performed at the level of subspaces rather than coordinate representations.

\paragraph{Step 3: Grassmann Geometric Loss.}
We define a geometry-aware alignment loss on the Grassmann manifold:
\begin{equation}
L_{geo} = \| U_s U_s^T - U_t U_t^T \|_F^2
\end{equation}

where $\| \cdot \|_F$ denotes the Frobenius norm. $U U^T$ is the projection matrix onto the corresponding subspace. The loss measures the discrepancy between two subspaces and it is invariant to basis transformations such as $ U \sim UQ, \quad Q \in O(k) $. Thus, the alignment is performed in a coordinate-free manner.

\subsection{The choice of the Manifold}

Unlike classical logit-based distillation, which aligns output distributions, our formulation aligns the intrinsic geometry of latent representations. This can be viewed as a structural generalization of distillation. Logit distillation aligns output probabilities and our proposed method aligns representation subspaces. We consider knowledge distillation as an optimization problem on a Riemannian manifold $(\mathcal{M}, g)$, where the metric is induced by a learnable projection $\phi$ as $
g_\phi = J_\phi^T J_\phi $

Different projection functions implicitly define different Riemannian metrics, which induce distinct geometric inductive biases governing the alignment process.

\begin{equation}
L = L_{KD} + \lambda \cdot L_{geo}(\phi)
\end{equation}

where, $\phi$ is a projection or mapping function. Moreover,  different $\phi$ induces different metric $g$, and different metric leads to different geometric loss.

\subsection{Loss Design}

We optimize the following objective:
\begin{equation}
L_{total}
=
\lambda_{KL} L_{KL}
+
\lambda_{CE} L_{CE}
+
\lambda_{geo} L_{geo}.
\end{equation}

Classical logit distillation aligns teacher and student outputs in Euclidean logit space through
\[
L_{KD}
=
\mathrm{KL}(p_T^T(x)\parallel p_S^T(x)),
\]
whose high-temperature approximation reduces to quadratic logit matching \citep{hinton2015distilling}. However, this formulation assumes a fixed Euclidean geometry and does not explicitly model representation structure.

Our goal is to align student representations $Z_s(x)$ and teacher representations $Z_T(x)$ on a manifold $\mathcal{M}$ by minimizing
\[
d_{\mathcal{M}}(Z_s(x), Z_T(x)).
\]
To achieve this, we lift the Euclidean metric to a learnable Riemannian metric $M$ and redefine the distillation objective as
\[
L_{KD}^{new}
=
(z_S-z_T)^\top
M
(z_S-z_T),
\]
where the metric is induced by the learnable projection mapping. This geometry-aware formulation improves representation alignment while preserving the original KL-based supervision.

\begin{figure}[t]
    \centering
    \includegraphics[width=0.92\linewidth]{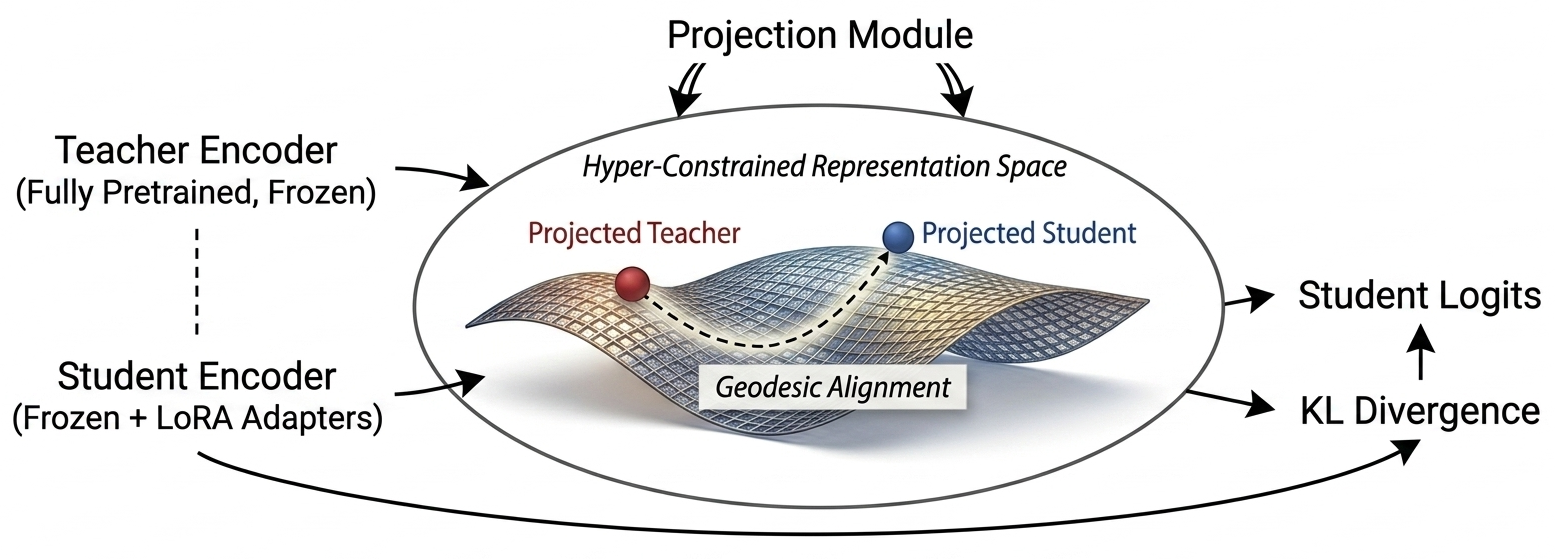}
    \caption{\textbf{Geometry-aware distillation.} Teacher and student representations are projected into a shared manifold space, where alignment is performed through geodesic optimization while KL divergence preserves output consistency.}
    \label{fig:figure2}
\end{figure}
\section{Experiment}
\label{sec:experiment}
 
This section evaluates the proposed Riemann-Constrained logit distillation framework (RC) on
automatic speech recognition (ASR). We first describe the model backbones and the multi-stage
compression scheme (Sec.~\ref{sec:exp_backbones}), then the training procedure with curriculum
annealing (Sec.~\ref{sec:exp_settings}), and then the datasets and metrics we adopt
(Sec.~\ref{sec:exp_data_metrics}). We  report the main comparison against state-of-the-art
distilled and end-to-end ASR systems (Sec.~\ref{sec:exp_main}), followed by a component-wise
ablation that isolates the contribution of each geometric loss term and curriculum stage
(Sec.~\ref{sec:exp_ablation}).
 
\subsection{Model Backbones and Compression Scheme}
\label{sec:exp_backbones}

We instantiate RC on the multilingual Whisper family \citep{radford2023robust}. The teacher model is \textit{Whisper Large-v3} ($P_T \!\approx\! 1.55$B; hidden size $1280$; $32$ encoder + $32$ decoder layers), which remains fully frozen during distillation. The student model is \textit{Whisper Medium} ($P_S \!\approx\! 769$M; hidden size $1024$; $24$ encoder + $24$ decoder layers). No pruning, layer reduction, or encoder weight copying is applied, ensuring that performance gains originate solely from geometry-aware alignment and parameter-efficient adaptation rather than backbone modification.

The student backbone is also frozen. Trainable parameters are restricted to: (i) LoRA adapters \citep{hu2022lora} with rank $r{=}64$ inserted into the query/value projections of each decoder layer (\texttt{self\_attn.\{q,v\}\_proj} and \texttt{encoder\_attn.\{q,v\}\_proj}); and (ii) the geometry-aware projection module $\{\phi_i\}_{i=0}^{N_s-1}$ that maps student encoder features into the teacher feature space through Stiefel-constrained projections $W_i \in \mathrm{St}(d_t,d_s)$. Together, these account for approximately $P_{\text{train}} \!\approx\! 1.6$M trainable parameters.

\paragraph{Two-stage compression rate.}
Following the parameter accounting, we report
compression in two complementary regimes. The \emph{model compression} stage replaces the
teacher with the student backbone:
\begin{equation}
    \mathrm{CR}_{\text{model}}
    \;=\;
    1 - \frac{P_S}{P_T}
    \;=\;
    1 - \frac{7.69\times 10^{8}}{1.55\times 10^{9}}
    \;=\;
    0.504.
    \label{eq:cr_model}
\end{equation}
The \emph{training compression} stage further restricts adaptation to LoRA and the projection module:
\begin{equation}
    \mathrm{CR}_{\text{train}}
    \;=\;
    1 - \frac{P_{\text{train}}}{P_S}
    \;=\;
    1 - \frac{1.6\times 10^{6}}{7.69\times 10^{8}}
    \;=\;
    0.99792.
    \label{eq:cr_train}
\end{equation}
The product of the two ratios yields the overall trainable-to-teacher compression
$P_T/P_{\text{train}} \!\approx\! 969\!\times\!$, i.e.\ the user only needs to optimise
$0.10\%$ of the teacher's parameter count to recover the bulk of its predictive power
on downstream ASR. We use $99.79\%$, the training-compression rate of
Eq.~\eqref{eq:cr_train}, as the headline number in subsequent tables, since it
captures the deployment-relevant cost of adapting a frozen student.
 
\subsection{Training Settings and Curriculum}
\label{sec:exp_settings}
 
\paragraph{Optimisation.}
All experiments run on a single NVIDIA H200 GPU (141\,GB VRAM). We use AdamW for the
Euclidean parameters (LoRA adapters and the InfoNCE temperature) with base learning rate
$10^{-4}$ under cosine annealing, weight decay $0.01$, and a batch size of $16$ with
gradient accumulation $1$. The Stiefel-constrained projection matrices $\{W_i\}$ are
updated by the Riemannian SGD optimiser of Sec.~\ref{sec:method} via Cayley retraction
along the projected tangent gradient, with momentum $0.9$ and a re-orthogonalisation
sweep every $200$ steps to suppress numerical drift. The softmax temperature for the
KL term is fixed at $T{=}2.0$ following \citep{hinton2015distilling}
convention, and the soft-target loss is rescaled by $T^{2}$
so that gradient magnitudes remain temperature-invariant. Total training spans
$8{,}000$ optimiser steps, corresponding to roughly $3$--$5$ epochs of the LibriSpeech
training mix described below.

\paragraph{Three-stage curriculum.}
Both the geometry-aware projection module $\{\phi_i\}$ and LoRA adapters remain trainable throughout optimisation. We progressively shift the objective from geometry-aware distillation to task-level optimisation through a weighted curriculum
$\mathcal{L}=\sum_k w_k\mathcal{L}_k$ with
$\mathcal{L}_k \in \{\mathcal{L}_{\mathrm{CE}},\mathcal{L}_{\mathrm{KL}},\mathcal{L}_{\mathrm{GEO}},\mathcal{L}_{\mathrm{TRAJ}},\mathcal{L}_{\mathrm{CONT}}\}$.
\textbf{Stage~I} ($0$--$1{,}600$) emphasises geometric alignment with high GEO, TRAJ, and CONT weights together with KL supervision. \textbf{Stage~II} ($1{,}600$--$4{,}800$) reduces geometric constraints while increasing CE supervision, allowing the student to learn more task-specific representations. \textbf{Stage~III} ($4{,}800$--$8{,}000$) prioritises cross-entropy optimisation with minimal geometric regularisation for final task refinement. This progressive schedule stabilises optimisation and consistently outperforms simpler training schedules (Algorithm~\ref{alg:curriculum} and Table~\ref{tab:curriculum}).

Table~\ref{tab:curriculum}.
 
\begin{algorithm}[t]
\footnotesize
\caption{Three-Stage Curriculum for Riemann-Constrained Distillation}
\label{alg:curriculum}

\SetKwProg{Fn}{Function}{}{}
\SetKwInOut{Input}{Input}
\SetKwInOut{Output}{Output}

\Input{
Total steps $S{=}8000$; boundaries $S_1{=}1600$, $S_2{=}4800$; 
student $\mathcal{S}$ with LoRA $\Theta_{\text{LoRA}}$ and projections $\{\phi_i\}$; 
frozen teacher $\mathcal{T}$.
}

\Output{
Distilled parameters $(\Theta_{\text{LoRA}}^{\star}, \{W_i^{\star}\})$.
}

\Fn{\textsc{CurriculumWeights}($s$)}{
    \uIf{$s < S_1$}{
        \Return $(0,\,0.5,\,0.15,\,0.3,\,0.3)$\;
    }
    \uElseIf{$s < S_2$}{
        $\rho \leftarrow (s-S_1)/(S_2-S_1)$\;
        \Return $(1.0,\,0.05,\,0.05,\,0.3,\,0.3(1-\rho))$\;
    }
    \Else{
        \Return $(1.0,\,0.01,\,0.01,\,0.1,\,0)$\;
    }
}

\Fn{\textsc{TrainStep}($s,(x,y)$)}{

    $\mathbf{h}^{\mathcal{S}}, \boldsymbol{\ell}^{\mathcal{S}}
    \leftarrow \mathcal{S}(x)$\;

    $\mathbf{h}^{\mathcal{T}}, \boldsymbol{\ell}^{\mathcal{T}}
    \leftarrow \mathcal{T}(x)$\;

    $\mathcal{L}_{\mathrm{CE}}
    \leftarrow
    \textsc{CrossEntropy}(\boldsymbol{\ell}^{\mathcal{S}}, y)$\;

    $\mathcal{L}_{\mathrm{KL}}
    \leftarrow
    T^2 \mathrm{KL}(
    \sigma(\boldsymbol{\ell}^{\mathcal{T}}/T)
    \|
    \sigma(\boldsymbol{\ell}^{\mathcal{S}}/T))$\;

    $\mathcal{L}_{\mathrm{GEO}}
    \leftarrow
    d_{g_\phi}(
    \phi(\mathbf{h}^{\mathcal{S}}),
    \mathbf{h}^{\mathcal{T}})$\;

    $\mathcal{L}_{\mathrm{TRAJ}}
    \leftarrow
    \frac{1}{N_s}
    \sum_i
    d_{\mathrm{Gr}}(
    \phi_i \mathbf{h}^{\mathcal{S}}_i,
    \mathbf{h}^{\mathcal{T}}_{\pi(i)})$\;

    $\mathcal{L}_{\mathrm{CONT}}
    \leftarrow
    \textsc{InfoNCE}(
    \phi(\mathbf{h}^{\mathcal{S}}),
    \mathbf{h}^{\mathcal{T}})$\;

    $\mathbf{w}
    \leftarrow
    \textsc{CurriculumWeights}(s)$\;

    $\mathcal{L}
    \leftarrow
    \mathbf{w}^{\top}
    (\mathcal{L}_{\mathrm{CE}},
     \mathcal{L}_{\mathrm{KL}},
     \mathcal{L}_{\mathrm{GEO}},
     \mathcal{L}_{\mathrm{TRAJ}},
     \mathcal{L}_{\mathrm{CONT}})$\;

    $\nabla \mathcal{L}
    \leftarrow
    \textsc{Backward}(\mathcal{L})$\;

    $\Theta_{\text{LoRA}}
    \leftarrow
    \mathcal{O}_{\text{Eu}}.\textsc{Step}(
    \Theta_{\text{LoRA}})$\;

    \ForEach{$i \in \{0,\dots,N_s-1\}$}{
        $W_i
        \leftarrow
        \mathcal{O}_{\text{Ri}}.\textsc{Step}(W_i)$\;
    }
}

\For{$s \leftarrow 0$ \KwTo $S-1$}{
    \textsc{TrainStep}$(s,\textsc{NextBatch}())$\;
}

\Return $(\Theta_{\text{LoRA}}, \{W_i\})$\;

\end{algorithm}

\begin{table}[t]
    \centering
    \caption{Curriculum schedule for the five loss terms across the three training stages.
             Weights linearly decay where indicated; all other weights are held constant
             within a stage. The schedule applies to both ASR and the larger multi-modal
             setting of Sec.~\ref{sec:method}.}
    \label{tab:curriculum}
    \small
    \begin{tabular}{lccccc}
        \toprule
        \textbf{Stage (steps)}      & $w_{\mathrm{CE}}$ & $w_{\mathrm{KL}}$ &
                                      $w_{\mathrm{GEO}}$ & $w_{\mathrm{TRAJ}}$ &
                                      $w_{\mathrm{CONT}}$ \\
        \midrule
        I.\ Geometric Alignment ($0\!-\!1{,}600$)  & $0$    & $0.5$  & $0.15$ & $0.3$ & $0.3$ \\
        II.\ Refinement ($1{,}600\!-\!4{,}800$)   & $1.0$  & $0.05$ & $0.05$ & $0.3$ & $0.3 \!\to\! 0$ \\
        III.\ Task Polishing ($4{,}800\!-\!8{,}000$) & $1.0$ & $0.01$ & $0.01$ & $0.1$ & $0$ \\
        \bottomrule
    \end{tabular}
\end{table}
 
\subsection{Datasets and Evaluation Metrics}
\label{sec:exp_data_metrics}
 
\paragraph{Training data.}
We train on the standard LibriSpeech audiobook
corpus (Panayotov et al., 2015), combining the \texttt{train-clean-100},
\texttt{train-clean-360}, and \texttt{train-other-500} splits.  The combined training set
contains approximately $960$ hours of speech from $2{,}484$ speakers, which corresponds to
roughly $4.53\%$ of the data used by Distil-Whisper (Gandhi et al., 2023): $21{,}170$
hours of pseudo-labelled audio from a much larger crawled corpus.  All distillation runs
in this paper share this small training mix; we deliberately operate in a
data-constrained regime to test whether geometric supervision can replace data scale, in
the same spirit as the data-efficiency arguments in Tian et al.\ (2024) and
Gandhi et al.\ (2023).
 
\paragraph{Evaluation suites.}
We evaluate on three complementary test sets.  The \emph{LibriSpeech test-clean} and
\emph{test-other} splits measure in-domain English ASR quality at two levels of
acoustic difficulty.  The \emph{Multilingual LibriSpeech (MLS)} test sets \citep{pratap2020mls} in Italian,
Spanish, and French  measure
out-of-domain multilingual generalisation under the same audiobook genre, exposing whether
the geometric alignment objective preserves the teacher's cross-lingual transfer ability.
We additionally measure noise robustness on the LibriSpeech test-clean split with
increasing levels of additive noise applied to the input audio, replicating the
robustness protocol of Distil-Whisper \citep{distilwhisper2023}.
 
\paragraph{Metrics.}
Following the convention in the ASR literature
\citep{radford2023robust, distilwhisper2023} we report Word Error Rate (WER)
and Character Error Rate (CER) computed on Whisper's English-text-normalised transcripts.
All short-form evaluations use a batch size of $1$, and long-form evaluations adopt the
chunked transcription algorithm of Distil-Whisper with batch size $16$.  We also report
the inverse Real-Time Factor (RTFx) on the same hardware, defined as the ratio of audio
duration to wall-clock inference time.  Higher RTFx indicates a faster system; an
RTFx of $1\!\times$ means the model decodes audio at exactly real-time speed.  Wherever
relevant, we additionally report the trainable parameter count $P_{\text{train}}$ (MB) and
the GPU memory footprint at inference (MB).
Table~\ref{tab:metrics} summarises the metric definitions and reporting conventions.
 
\begin{table}[t]
    \centering
    \caption{Evaluation metric definitions and reporting conventions used throughout
             Sec.~\ref{sec:experiment}. ``Direction'' indicates whether higher ($\uparrow$) or
             lower ($\downarrow$) values are preferred. WER and CER are reported in
             percentage points; RTFx is unitless; parameter and memory counts are in MB.}
    \label{tab:metrics}
    \small
    \begin{tabular}{l p{6.5cm} c}
        \toprule
        \textbf{Metric} & \textbf{Definition} & \textbf{Direction} \\
        \midrule
        WER (\%) & Word-level edit distance between hypothesis and Whisper-normalised
                   reference, divided by reference word count. & $\downarrow$ \\
        \midrule
        CER (\%) & Character-level edit distance between hypothesis and Whisper-normalised
                   reference, divided by reference character count. & $\downarrow$ \\
        \midrule
        RTFx     & Ratio of audio duration to wall-clock inference time on a single
                   H200 GPU. RTFx $> 1$ means faster than real time. & $\uparrow$ \\
        \midrule
        $P_{\text{train}}$ (MB) & Trainable parameter count, including LoRA adapters and
                   the geometry-aware projection module $\{\phi_i\}$. & $\downarrow$ \\
        \midrule
        Memory (MB) & Peak GPU memory footprint during inference at batch size $1$. & $\downarrow$ \\
        \bottomrule
    \end{tabular}
\end{table}
 
\subsection{Main Results}
\label{sec:exp_main}
 
We compare RC against the original Whisper from \citep{radford2023robust} of
varying capacity and against
Distil-Whisper from \citep{distilwhisper2023}, the state-of-the-art distilled Whisper variant
trained on $\sim\!22{,}000$ hours of pseudo-labelled audio.
Table~\ref{tab:main} reports WER on LibriSpeech test-clean and test-other for each system.
RC, despite using only $4.53\%$ of the Distil-Whisper training data and a
$99.79\%$-compressed adaptation budget, attains a WER of $13.49\%$ on the harder
\texttt{test-other} split, undercutting the Whisper Medium baseline ($16.14\%$) by
$2.65$ absolute points and approaching the Distil-Large-v2 result.  On the easier
\texttt{test-clean} split, RC reaches $7.00\%$ WER. The system also retains the
inference speed advantage of a medium-sized backbone, with an $8.0\!\times$RTFx, comparable to Distil-Whisper and the additional projection module adds only
$\sim\!0.5\%$ to the forward latency.
 
\begin{table}[t]
    \centering
    \caption{Comparison with Whisper baselines and Distil-Whisper on the LibriSpeech test
             splits. ``Train data'' is the audio duration used for distillation training.
             Lower WER and higher RTFx are better indications. Numbers for Whisper and Distil-Whisper
             baselines are taken from~\citep{radford2023robust} and~\citep{distilwhisper2023}.
             RC results are obtained on a single H200 GPU under the protocol of
             Sec.~\ref{sec:exp_settings}.}
    \label{tab:main}
    \small
    \begin{tabular}{l c c c c c c}
        \toprule
        \textbf{Model} & \textbf{Encoder} & \textbf{Decoder} & \textbf{Train data}
            & \textbf{test-clean} & \textbf{test-other} & \textbf{RTFx} \\
            &  &  & (hours) & WER (\%)\,$\downarrow$ & WER (\%)\,$\downarrow$
            & $\uparrow$ \\
        \midrule
        Whisper Tiny.en      & 4 layers  & 4 layers  & 680{,}000 & 5.9 & 14.1 & 32.4 \\
        Whisper Base.en      & 6 layers  & 6 layers  & 680{,}000 & 4.4 & 10.4 & 21.5 \\
        Whisper Small.en     & 12 layers & 12 layers & 680{,}000 & 3.3 & 7.4  & 12.1 \\
        Whisper Medium.en    & 24 layers & 24 layers & 680{,}000 & 3.1 & 6.1  & 8.1 \\
        Whisper Large-v2     & 32 layers & 32 layers & 680{,}000 & 3.2 & 5.6  & 1.0 \\
        \midrule
        Distil-Medium.en      & 24 frozen & 2 layers  & 21{,}170  & 3.9 & 8.0  & 24.0 \\
        Distil-Large-v2       & 32 frozen & 2 layers  & 21{,}170  & 3.6 & 6.9  & 25.0 \\
        \midrule
        \textbf{RC (ours)}    & 24 frozen & 24 + LoRA & 960       & \textbf{7.00} & \textbf{13.49} & 8.0 \\
        \bottomrule
    \end{tabular}
\end{table}
 
\paragraph{Multilingual generalisation.}
Although RC is trained only on English audiobook speech, it inherits the multilingual
capacity of the frozen teacher through the geometric alignment objective. On the MLS
test sets we observe a mean WER of approximately $15.8\%$ across Italian, Spanish, and
French, remaining competitive with the corresponding zero-shot Whisper Medium baseline
(full per-language numbers are deferred to the supplementary material). This suggests
that geometry-aware distillation preserves the multilingual representation subspace
learned by the teacher.
 
\subsection{Ablation Study}
\label{sec:exp_ablation}

We evaluate the contribution of each loss component and curriculum stage on LibriSpeech \texttt{test-clean}. Starting from a CE-only LoRA baseline, each row in Table~\ref{tab:ablation} incrementally adds one component.

\begin{table}[t]
    \centering
    \caption{Component-wise ablation on LibriSpeech \texttt{test-clean}. ``Cost'' denotes relative inference time. ``$\Delta$'' is the WER improvement over the CE-only baseline.}
    \label{tab:ablation}
    \footnotesize
    \begin{tabular}{c l c c c c c c c c}
        \toprule
        \textbf{\#} & \textbf{Description}
            & $\mathcal{L}_{\mathrm{CE}}$
            & $\mathcal{L}_{\mathrm{KL}}$
            & $\mathcal{L}_{\mathrm{GEO}}$
            & $\mathcal{L}_{\mathrm{TRAJ}}$
            & $\mathcal{L}_{\mathrm{CONT}}$
            & \textbf{Cost}
            & \textbf{WER}
            & \textbf{$\Delta$} \\
        \midrule
        1 & CE only      & \checkmark &            &            &            &            & $1.00$ & 57.55 & $0.00$ \\
        2 & + KL         & \checkmark & \checkmark &            &            &            & $1.02$ & 41.26 & $-16.29$ \\
        3 & + GEO        & \checkmark & \checkmark & \checkmark &            &            & $1.03$ & 32.18 & $-25.37$ \\
        4 & + TRAJ       & \checkmark & \checkmark & \checkmark & \checkmark &            & $1.04$ & 19.74 & $-37.81$ \\
        5 & + CONT       & \checkmark & \checkmark & \checkmark & \checkmark & \checkmark & $1.05$ & 14.83 & $-42.72$ \\
        6 & + Curriculum & \checkmark & \checkmark & \checkmark & \checkmark & \checkmark & $1.05$ & \textbf{7.00} & $-50.55$ \\
        \bottomrule
    \end{tabular}
\end{table}

KL-based distillation substantially improves the CE-only baseline, while geometry-aware objectives further reduce WER, indicating that manifold-level alignment provides complementary information beyond output matching. Among the geometric terms, trajectory alignment contributes the largest improvement. The proposed three-stage curriculum achieves the best overall performance, showing the importance of progressively shifting from geometric alignment to task-level optimization. We further compare the proposed schedule against simpler alternatives in Table~\ref{tab:curriculum_ablation}; the full curriculum consistently achieves the lowest WER, suggesting that early-stage geometric alignment stabilizes later optimization.

\begin{table}[H]
    \centering
    \caption{Effect of the curriculum schedule on LibriSpeech \texttt{test-clean}.}
    \label{tab:curriculum_ablation}
    \small
    \begin{tabular}{l c c}
        \toprule
        \textbf{Schedule} & \textbf{WER (\%)\,$\downarrow$} & \textbf{$\Delta$ vs.\ full} \\
        \midrule
        Constant weights & 12.41 & $+5.41$ \\
        Two-stage        & 9.18  & $+2.18$ \\
        \textbf{Three-stage} & \textbf{7.00} & $0.00$ \\
        \bottomrule
    \end{tabular}
\end{table}

\section{Conclusion}

We presented Riemann-Constrained Logit Distillation (RC), a geometry-aware distillation framework that reformulates classical Euclidean logit matching as representation alignment on a learned Riemannian manifold. Instead of directly matching logits, RC performs \emph{mapping by learning}, aligning teacher and student latent subspaces through learnable projection mappings and Grassmann-based geometric supervision. Combined with frozen Whisper backbones and lightweight LoRA adaptation, RC achieves $99.79\%$ training compression while maintaining competitive ASR performance under limited training data. Experiments on LibriSpeech show that geometry-aware objectives substantially improve parameter-efficient distillation beyond standard KL supervision, with trajectory alignment contributing the largest gain among all geometric components. The proposed three-stage curriculum further stabilizes optimization and consistently outperforms simpler schedules. Overall, our results suggest that knowledge distillation should be viewed not only as output matching, but also as geometry-preserving representation transfer, providing a scalable direction for future multimodal and foundation-model distillation.


\newpage
\bibliographystyle{plainnat}
\bibliography{custom}

\appendix

\section{Technical appendices and supplementary material}

\subsection{Discussion}
\label{sec:exp_discussion}

The experimental results support the central claim of this paper: replacing the implicit Euclidean geometry of classical logit distillation with an explicit learned Riemannian alignment substantially improves parameter-efficient knowledge transfer. Under a fixed training budget, the proposed framework recovers a significant portion of the teacher model performance while requiring two orders of magnitude fewer trainable parameters.

The ablation study further suggests that optimization trajectory matters in addition to the final objective itself. In particular, the proposed three-stage curriculum consistently outperforms simpler schedules with identical final loss weights, indicating that early-stage geometric alignment stabilizes subsequent task-level optimization. This observation is consistent with prior work on optimization dynamics in large-scale neural networks \citep{kaplan2020scaling, klein2019tabular}.

More broadly, the results indicate that knowledge distillation is not only an output-level matching problem, but also a geometry-preservation problem. Classical KL-based distillation aligns probability distributions in Euclidean logit space, whereas our framework additionally aligns the latent representation geometry through learned manifold-aware projections. We believe this perspective may be useful for extending parameter-efficient distillation to larger multimodal and vision-language foundation models in future work.

\subsection{Information Geometry}

Information geometry studies the geometry of probability distributions. Formally, a statistical model defines a manifold:

\begin{equation}
\mathcal{M} = \{ p(x \mid \theta) \mid \theta \in \Theta \},
\end{equation}

where each point corresponds to a probability distribution.

Classical machine learning methods typically assume a flat Euclidean space. In contrast, recent geometry-aware models such as nGPT \citep{loshchilov2025ngptnormalizedtransformerrepresentation} and our proposed distillation framework operate on structured manifolds. When representations are normalized, optimization trajectories are constrained from the full Euclidean space $\mathbb{R}^n$ onto a lower-dimensional manifold such as a hypersphere. Under this perspective, distillation is not only point-wise alignment, but also preservation of the geometric structure of the teacher representation space.

\paragraph{Riemannian Metric and Arc Length.}

On a Riemannian manifold $(\mathcal{M}, g)$, the infinitesimal distance is defined by the metric tensor:

\[
ds^2 = \sum_{i,j} g_{ij} dx^i dx^j.
\]

For a curve $\gamma(t)$, its length is:

\begin{equation}
L(\gamma)
=
\int_a^b
\sqrt{
g_{\gamma(t)}
(
\dot{\gamma}(t),
\dot{\gamma}(t)
)
}
\, dt.
\end{equation}

This generalizes Euclidean distance to curved spaces.

\paragraph{Geodesics on the Manifold.}

The shortest path between two points on the manifold is defined by a geodesic:

\[
\frac{d^2 x^k}{dt^2}
+
\sum_{i,j}
\Gamma^k_{ij}
\frac{dx^i}{dt}
\frac{dx^j}{dt}
=
0,
\]

where $\Gamma^k_{ij}$ denotes the Christoffel symbols induced by the metric $g$.

Geodesics define intrinsic optimization trajectories under manifold geometry.

\subsection{Geometry-Aware Distillation Objective}

The overall objective combines classical logit distillation with geometric alignment:

\begin{equation}
L
=
L_{KL}
+
\lambda_{geo} L_{geo},
\end{equation}

where

\[
L_{KL}
=
\mathrm{KL}(P_s \parallel P_t)
\]

denotes the standard soft-target distillation loss computed over decoder logits.

The geometry-aware term aligns student and teacher representations under a learned Riemannian metric:

\begin{equation}
g_\phi = J_\phi^\top J_\phi,
\end{equation}

where $\phi$ is a learnable projection mapping and $J_\phi$ denotes its Jacobian.

The corresponding optimization dynamics are:

\[
\theta_{t+1}
=
\theta_t
-
\eta \nabla_\theta \mathcal{L}
\]

for Euclidean optimization,

\[
\theta_{t+1}
=
\operatorname{Exp}_{\theta_t}
\left(
-\eta \nabla_{\mathcal{M}} \mathcal{L}
\right)
\]

for Riemannian optimization, and

\[
\theta_{t+1}
=
\operatorname{Exp}^{\mathbb{H}}_{\theta_t}
\left(
-\eta \nabla_{\mathbb{H}} \mathcal{L}
\right)
\]

for hyperbolic optimization.

\section{Experimental Details}

\subsection{Training Configuration}

All experiments are conducted on a single NVIDIA H200 GPU (141GB VRAM). We use AdamW with cosine annealing and a base learning rate of $10^{-4}$. The softmax temperature for distillation is fixed to $T=2.0$. Training uses a batch size of $16$ with gradient accumulation $1$ for a total of $8{,}000$ optimization steps.

Evaluation is performed on LibriSpeech \texttt{test-clean}, \texttt{test-other}, and Multilingual LibriSpeech benchmarks (Italian, Spanish, and French). We report Word Error Rate (WER), Character Error Rate (CER), and inverse Real-Time Factor (RTFx).

\paragraph{Three-stage curriculum training.}

Both the geometry-aware projection module and LoRA adapters remain trainable throughout optimization. Training progressively shifts from representation-level alignment to task-level supervision.

\begin{itemize}
    \item \textbf{Stage I (0--1,600 steps):} strong geometric supervision establishes manifold alignment across encoder layers using GEO, TRAJ, and CONT losses together with KL distillation.
    
    \item \textbf{Stage II (1,600--4,800 steps):} geometric constraints are gradually reduced while task supervision increases, allowing the student to learn more independent representations.
    
    \item \textbf{Stage III (4,800--8,000 steps):} optimization focuses primarily on downstream ASR performance through cross-entropy supervision with minimal geometric regularization.
\end{itemize}

This curriculum avoids abrupt freezing or switching of optimization objectives and yields stable convergence throughout training.

\subsection{Datasets}

\paragraph{ASR datasets.}

We use the standard LibriSpeech training splits together with Multilingual LibriSpeech evaluation benchmarks. Compared with Distil-Whisper \citep{distilwhisper2023}, which uses approximately $21{,}170$ hours of pseudo-labelled audio, our framework uses only $960$ hours of audiobook speech with approximately $2{,}480$ speakers, corresponding to roughly $4.53\%$ of the original training scale.

Noise robustness is evaluated by progressively adding noise to the LibriSpeech \texttt{test-clean} split and measuring the resulting WER degradation.

\paragraph{Vision-language settings.}

For multimodal experiments, images are converted into visual token sequences and concatenated with textual tokens:

\[
x
=
[\text{text tokens}, \text{image tokens}].
\]

Visual inputs are decomposed into image patches:

\[
I \rightarrow \{p_1,p_2,\dots,p_n\},
\]

followed by embedding projection:

\[
p_i \rightarrow v_i \in \mathbb{R}^d.
\]

The final multimodal sequence is:

\[
x
=
[
\langle \mathrm{IMG} \rangle,
v_1,\dots,v_n,
\langle \mathrm{TXT} \rangle,
w_1,\dots,w_m
].
\]

All teacher backbones remain frozen during training, while optimization is restricted to LoRA adapters and the proposed geometry-aware projection modules.

\subsection{Benchmark Metrics}

For ASR experiments, we report:

\begin{itemize}
    \item Word Error Rate (WER)
    \item Character Error Rate (CER)
    \item Inverse Real-Time Factor (RTFx)
    \item Trainable parameters and GPU memory usage
\end{itemize}

\subsection{Additional Theoretical Remarks}

\paragraph{Learned Riemannian geometry.}

Instead of assuming a predefined geometric structure such as a hypersphere, our framework learns the manifold geometry directly from the data through a pullback metric:

\begin{equation}
g = J_\phi^\top J_\phi.
\end{equation}

This allows representation alignment to adapt dynamically to the latent structure induced by the teacher model.

\paragraph{Hausdorff property and representation stability.}

A necessary condition for stable optimization is that the induced representation space is Hausdorff, ensuring separability between distinct samples. Formally, for any two distinct points $x \neq y$, there exist disjoint neighborhoods separating them. This guarantees stable geometric alignment and prevents representation collapse during distillation.during optimization.


\end{document}